\documentclass{article}

\PassOptionsToPackage{numbers}{natbib}

\usepackage[preprint]{neurips_2026}

\usepackage[utf8]{inputenc} 
\usepackage[T1]{fontenc}    
\usepackage{hyperref}       
\usepackage{url}            
\usepackage{booktabs}       
\usepackage{amsfonts}       
\usepackage{nicefrac}       
\usepackage{microtype}      
\usepackage{xcolor}         
\usepackage{multirow}
\usepackage{graphicx}
\usepackage{subcaption}
\usepackage[table]{xcolor}
\usepackage{amsmath}
\usepackage{amssymb}
\usepackage{wrapfig}
\usepackage{enumitem}
\usepackage{algorithm}
\usepackage{algorithmic}
\usepackage{adjustbox}
\usepackage{caption}
\newcommand{\eg}{\textit{e}.\textit{g}.}
\newcommand{\ie}{\textit{i}.\textit{e}.}

\newcounter{bxincomm}
\definecolor{aqua}{rgb}{0.00,0.67,0.80}

\newcounter{todocomm}

\hypersetup{colorlinks=true,linkcolor=red!70!black,linktocpage=false,citebordercolor=blue!70!black,citecolor=blue!70!black,anchorcolor=blue!70!black}

\title{Protein Functional Semantic Alignment via Representation Rectification}

\author{
  \textbf{Chuyang Zhou}$^{1*}$,
  \textbf{Bingxin Zhou}$^{2*}$,
  \textbf{Andi Han}$^{1}$,
  \textbf{Chang Xu}$^{1}$ \\[0.5em]
  $^{1}$The University of Sydney \\
  $^{2}$Shanghai Jiao Tong University \\[0.3em]
  \texttt{\{chuyang.zhou,andi.han,c.xu\}@sydney.edu.au} \\
  \texttt{bingxin.zhou@sjtu.edu.cn}
}

\begin{document}

\maketitle

\begingroup
\renewcommand{\thefootnote}{*}
\footnotetext{Equal Contribution.}
\endgroup

\begin{abstract}
  Protein language models (PLMs) are commonly pretrained with likelihood-based objectives such as masked language modeling or denoising. These objectives capture sequence-level regularities, but can produce representations in which functional signals are entangled with non-functional variation (\eg, sequence identity, compositional bias, and frequent local motifs), creating a readout bottleneck for downstream function prediction. We introduce Functional \textbf{S}emantic \textbf{A}lignment \textbf{v}ia Representation \textbf{R}ectification (SAVR), a contrastive post-training method using Pfam family co-membership as a homology-guided discrete semantic signal to reshape PLM representations. SAVR reduces directional dominance and redistributes variance across embedding dimensions, decreasing variance concentration, and increasing effective dimensionality of the embedding covariance spectrum. Without observing any task-specific labels during post-training, SAVR achieves state-of-the-art results on Enzyme Commission (EC) benchmarks and competitive results on Gene Ontology (GO) benchmarks under frozen-backbone evaluation. These results suggest that homology-guided contrastive post-training can rectify PLM representation geometry and improve sequence-based protein function prediction. 
\end{abstract}

\section{Introduction}

\begin{figure*}[ht]
\vskip -0.1in
  \begin{center}
  \centering
    \includegraphics[width=\textwidth]{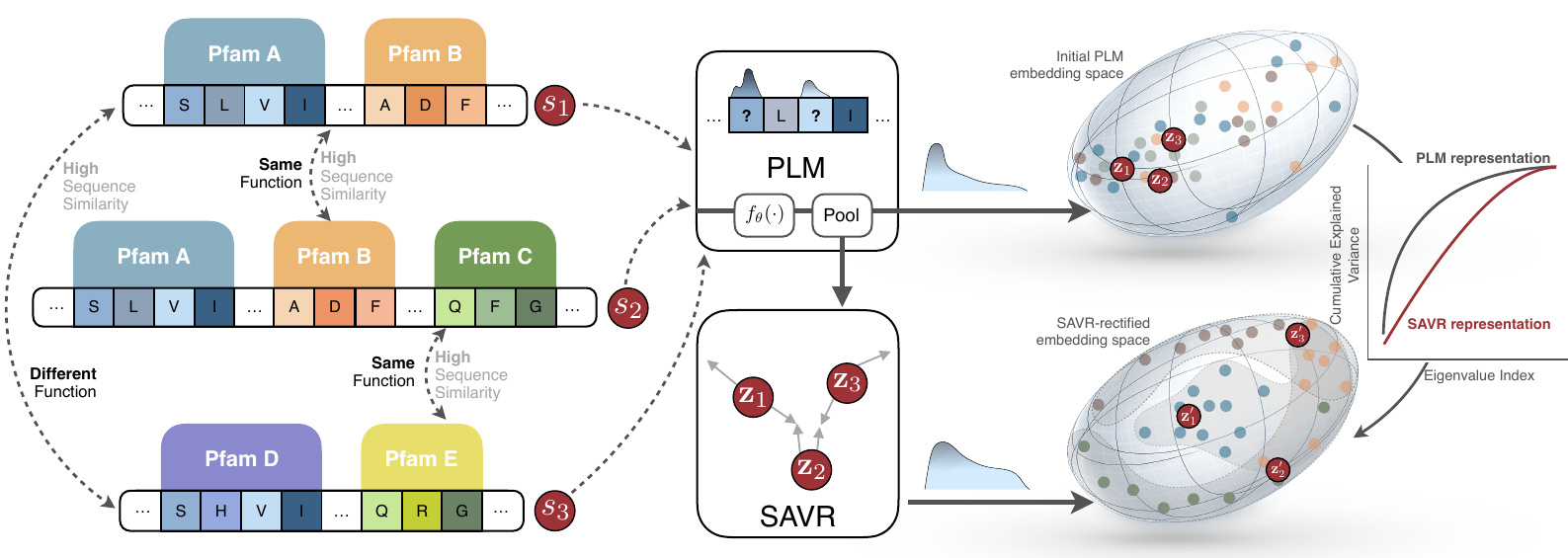}
    \caption{\textbf{Overview of SAVR}. SAVR reshapes PLM representations using family co-membership as a homology-guided contrastive signal. \textbf{Left}: raw sequence similarity does not always provide reliable positive pairs, since locally similar proteins may differ in domain composition and function, whereas proteins with lower local similarity may still share family-level homology. \textbf{Middle}: family co-membership defines contrastive positive pairs, pulling together homology-related sequences while contrasting them against negatives. \textbf{Right}: post-training reshapes the embedding space from an elongated, mixed point cloud to a rounder space with clearer grouping, and makes the eigenvalue spectrum less steep, indicating reduced variance concentration across directions.
    }
    \label{intro-fig}
  \end{center}
\end{figure*}

Protein language models (PLMs) have emerged as a central paradigm in computational biology, demonstrating strong capability in capturing biological regularities from large-scale unlabeled protein sequences~\cite{elnaggar2021prottrans,lin2023evolutionary,zhou2024mlife}. Trained with self-supervised objectives such as masked language modeling, PLMs learn high-dimensional representations that support a broad range of downstream applications, such as structure prediction~\cite{jumper2021highly, baek2021accurate, lin2023evolutionary} and function prediction related tasks such as Enzyme Commission prediction~\cite{yu2023enzyme, yang2025interpretable} and Gene Ontology prediction~\cite{radivojac2013large, jiang2016expanded, zhou2019cafa}.

Recent progress in representation analysis suggests that model capability depends not only on \emph{what} information is encoded, but also on \emph{how} that information is organized in embedding space~\cite{li2025tracing}. In particular, geometric diagnostics such as anisotropy and spectral concentration have been used to characterize when representations are dominated by a small number of directions, which can affect the behavior of readout models~\cite{, ethayarajh2019contextual, garrido2023rankme, huang2021whiteningbert}.

Motivated by this perspective, we revisit a practical limitation of pretrained PLM backbones for function prediction. Token-level likelihood objectives encourage encoding statistically predictable patterns in sequences, which include factors that are not directly aligned with functional semantics~\cite{bepler2021learning,brandes2022proteinbert}. As a result, the protein sequence representations can entangle functional signals with non-functional variation such as sequence identity, compositional bias, and frequent local motifs~\cite{lin2023evolutionary,marquet2022embeddings}. This entanglement is often reflected in a compressed and anisotropic embedding geometry, where a few dominant directions explain a disproportionate variance, making function readout less reliable.



In this work, we propose to rectify PLM representations using a discrete intermediate semantic signal, without altering the pretraining objective or relying on task-specific labels. 
We introduce \textbf{Functional Semantic Alignment via Representation Rectification (SAVR)}, a contrastive post-training method that uses Pfam~\cite{bateman2000pfam}, a curated collection of protein families and domains, and treats Pfam co-membership as a weak semantic relation to form positive pairs at the sequence level. 
SAVR brings together sequences that share Pfam annotations, capturing domain-level semantics, while pushing apart sequences from different Pfam groups, thereby reshaping the representation space toward function-relevant organization.

From a geometric viewpoint, as illustrated in Figure~\ref{intro-fig}, SAVR reduces directional dominance and redistributes variance across embedding dimensions, decreasing variance concentration and increasing the effective dimensionality of the embedding covariance spectrum. We implement SAVR with parameter-efficient low-rank adaptation, enabling lightweight post-training while keeping the base encoder architecture unchanged. We show that on different PLM backbone architectures, SAVR consistently reduces spectral concentration and anisotropy in embeddings (\ie, lower variance concentration in leading principal directions and higher effective rank, as shown in Section~\ref{sec:geom}). Empirically, these representation-level changes lead to state-of-the-art results on Enzyme Commission (EC) benchmarks and competitive Gene Ontology (GO) performance, with clear gains under comparable single-sequence input settings.

In summary, our contributions are threefold.
First, \textbf{we reveal geometry-level readout bottlenecks in pretrained PLM embeddings}. With spectral concentration, effective rank, and anisotropy-related metrics, we show that function-relevant signals can be difficult to extract when embeddings are dominated by a few principal directions (Section~\ref{sec:formulation}).
Second, \textbf{we propose SAVR to reshape PLM representations through parameter-efficient contrastive post-training}. SAVR constructs homology-guided sequence pairs without explicitly using EC/GO labels during post-training (Section~\ref{sec:method}).
Third, \textbf{we show that representation rectification improves protein function prediction}. SAVR consistently rectifies embedding geometry across PLM backbones, achieves state-of-the-art EC results, and remains competitive on GO under single-sequence inference (Section~\ref{sec:exp}).


\section{Geometry-Level Readout Bottlenecks in PLM Representations}
\label{sec:formulation}
We study post-training strategies that improve protein function prediction by rectifying the geometric structure of pretrained protein representations.

\subsection{Protein Sequence Representations}
\label{sec:repr}
Let $\mathcal{S}$ denote the space of protein sequences. A sequence $s=(x_1,\ldots,x_L)\in\mathcal{S}$ is a length-$L$ amino-acid token sequence. We consider a PLM encoder $f_\theta(\cdot)$ ($\theta$ represents learnable parameters) that maps $s$ to contextualized token embeddings of hidden size $d$: $\mathbf{H}_s=f_\theta(s)\in\mathbb{R}^{L\times d}.$
To obtain a fixed-dimensional representation for the full protein, we apply a pooling operator over non-padding, non-special tokens (mean pooling in our implementation) to produce $\mathbf{z_s}=\mathrm{Pool}(\mathbf{H}_s)\in\mathbb{R}^{d}.$

Our focus is the geometry of pooled representations $\{\mathbf{z}_s\}$ induced by likelihood-based pretraining. We analyze this geometry because downstream function predictors typically operate on fixed protein-level embeddings, and their performance can depend not only on whether functional information is encoded, but also on whether that information is accessible to a simple readout. In particular, when representation variance is concentrated along a few dominant directions, function-relevant variation may be entangled with nuisance factors and become harder to separate with lightweight predictors. We therefore quantify this geometry using covariance-spectrum diagnostics in Section~\ref{sec:geo_metrics}.

\subsection{Covariance-Spectrum Diagnostics of Representation Geometry}
\label{sec:geo_metrics}
To characterize geometry-level readout bottlenecks, we analyze the covariance spectrum of pooled protein embeddings. A concentrated spectrum indicates that representation variance is dominated by a small number of directions, which can make function-relevant variation less accessible to lightweight downstream predictors. We use three complementary diagnostics: (1) \textbf{Top-$k$ ratio} for directional dominance, (2) \textbf{RankMe} for effective dimensionality, and (3) \textbf{spectral and piecewise decay exponents} for global and local spectrum shape. Empirical analysis are in Section~\ref{sec:geom}.


We first define the basic setup. 
Given a set of $N$ pooled representations $\{\mathbf{z}_i\}_{i=1}^{N}$, let $\tilde{\mathbf{Z}}\in\mathbb{R}^{N\times d}$ denote the centered representation matrix, whose rows are
$\tilde{\mathbf{z}}_i=\mathbf{z}_i-\frac{1}{N}\sum_{j=1}^{N}\mathbf{z}_j$.
The empirical covariance matrix is
\begin{equation}
    \boldsymbol{\Sigma}=\frac{1}{N-1}\tilde{\mathbf{Z}}^\top \tilde{\mathbf{Z}}\in\mathbb{R}^{d\times d},
\end{equation}
and its eigenvalues are sorted as $\lambda_1\geq\cdots\geq\lambda_d\geq 0$.

\textbf{Effective rank (RankMe).}
We use RankMe~\cite{garrido2023rankme} as an effective rank measure based on spectral entropy. Define normalized eigenvalues
$p_i=\frac{\lambda_i}{\sum_j\lambda_j+\epsilon}$, where $\epsilon$ is a small constant for numerical stability.
RankMe is defined as
\begin{equation}
    \mathrm{RankMe}=\exp\!\left(-\sum_{i=1}^{d} p_i \log(p_i+\epsilon)\right).
\end{equation}
Higher RankMe indicates variance is distributed more uniformly across embedding dimensions. 

\textbf{Spectral concentration (Top-$k$ ratio).}
As a simple proxy for anisotropy, we report the fraction of total variance explained by the Top-$k$ principal directions:
\begin{equation}
    \text{Top-}k=\frac{\sum_{i=1}^{k}\lambda_i}{\sum_{j=1}^{d}\lambda_j+\epsilon}.
\end{equation}
Higher Top-$k$ indicates stronger variance concentration. We report Top-$10$ by default as variance concentration in PLMs is typically dominated by the leading directions.

\textbf{Spectral decay exponent ($\alpha_{\mathrm{req}}$).}
To summarize spectral decay beyond Top-$k$ and RankMe, we fit a power-law slope~\cite{agrawal2022alreq} to the full covariance spectrum $\{\lambda_i\}_{i=1}^d$.
Specifically, we perform a least-square fit in log--log space:
\begin{equation}
    \log(\lambda_i+\epsilon) = -\alpha_{\mathrm{req}} \log i + c, \quad i=1,\ldots,d,
\end{equation}
and report $\alpha_{\mathrm{req}}$ together with $R^2$ as a goodness-of-fit indicator.
$\alpha_{\mathrm{req}}$ can be used as a descriptor of spectral decay.

\textbf{Piecewise decay exponents ($\alpha_h,\alpha_m,\alpha_t$).}
A single global slope can obscure different behaviors between the leading, middle, and tailing parts of the spectrum. We therefore fit piecewise exponents on three segments defined by cumulative explained variance. Let
\begin{equation}
    C(k)=\frac{\sum_{i=1}^{k}\lambda_i}{\sum_{j=1}^{d}\lambda_j+\epsilon}.
\end{equation}
With fixed thresholds $\rho_a=0.4$ and $\rho_b=0.8$, define $k_t=\min\{k\in[1,d]:C(k)\geq\rho_t\},\quad t\in\{a,b\}.$
This gives three spectrum segments: head $[1,k_a]$, middle $[k_a+1,k_b]$, and tail $[k_b+1,d]$. We fit $\log(\lambda_i+\epsilon)=-\alpha\log i+c$ on each segment and report $(\alpha_h,\alpha_m,\alpha_t)$. This variance-based segmentation adapts the index ranges to each representation while fixing protocol across models.



\section{SAVR: Homology-Guided Representation Rectification}
\label{sec:method}
Having diagnosed geometry-level readout bottlenecks in pretrained PLM embeddings, we next introduce Functional \textbf{S}emantic \textbf{A}lignment \textbf{v}ia Representation \textbf{R}ectification (\textbf{SAVR}), a contrastive post-training method that reshapes representation geometry before downstream task-specific training. SAVR constructs homology-guided sequence pairs from family co-membership and post-trains a PLM backbone with parameter-efficient low-rank adapters. It encourages homology-related sequences to have nearby representations while maintaining separation from negatives. After post-training, the refined PLM produces single-sequence embeddings and can be used with lightweight downstream predictors without requiring retrieval or external annotations at inference time.

\subsection{Parameter-Efficient Backbone Adaptation}
\label{sec:method_backbone}
Let $f_\theta$ denote a pretrained PLM encoder and $\mathbf{z}(s)\in\mathbb{R}^d$ the pooled sequence representation from Section~\ref{sec:repr}. SAVR freezes the pretrained backbone parameters $\theta$ and inserts LoRA~\cite{hu2022lora} adapters with trainable parameters $\Delta\theta$ into the attention layers. During post-training, only $\Delta\theta$ and a lightweight projection head are updated. The resulting adapted encoder is denoted by $f_{\theta+\Delta\theta}$.

\subsection{Homology-Guided Positive Pair Construction}
\label{sec:pfam}
SAVR uses protein family co-membership to construct homology-guided positive pairs for contrastive post-training. In practice, we use Pfam~\cite{bateman2000pfam}, a curated database of protein families and domains, to define a binary relation. This homology-guided rule plays a role similar to sequence-similarity thresholding, but avoids manually selecting a fixed identity cutoff across heterogeneous protein families. Specifically, two sequences are treated as a positive pair if they share at least one annotated family. This binary relation is intentionally simple and applicable: in our post-training data, $77.4\%$ of proteins have exactly one annotated Pfam domain type and $93.6\%$ have at most two, so most pair assignments are not dominated by complex multi-domain ambiguity. For the same reason, we do not use Pfam hierarchies or annotation-confidence weights during pair construction.

Importantly, family co-membership is used only to construct training-time pair relations. SAVR does \textit{not} optimize a family-classification objective and does \textit{not} explicitly use Pfam annotations during post-training. After post-training, downstream prediction remains single-sequence: the adapted encoder produces embeddings directly from the input sequence, without retrieval or external annotations at inference time. This design provides a scalable middle ground between raw sequence similarity and endpoint functional labels. Compared with fixed local sequence-identity thresholds, family co-membership captures broader homology relations that can remain useful across heterogeneous protein families. At the same time, it is not tied to any specific downstream label space, making it suitable as a general post-training relation for protein-level function prediction.

Formally, let $\mathcal{F}(s)$ be the set of protein families associated with sequence $s$. For two sequences $s_i$ and $s_j$, we define them as a positive pair if they share at least one family:
\begin{equation}
    \mathcal{F}(s_i)\cap\mathcal{F}(s_j)\neq\emptyset.
\end{equation}
This binary relation is intentionally coarse. It does not imply functional equivalence between two proteins, but provides a scalable homology-oriented relation for contrastive post-training. 

Within a contrastive batch $\mathcal{B}$, the in-batch positive set for anchor $i$ is
\begin{equation}
\mathcal{P}_{\mathcal{B}}(i)=\{\, j\in \mathcal{B}\setminus\{i\}\mid F(s_i)\cap F(s_j)\neq \emptyset \,\}.
\end{equation}
To ensure that each batch contains positive examples, we use a family-bucketed sampler. We maintain buckets $\{\mathcal{B}_p\}_{p\in\mathcal{P}}$, where $\mathcal{B}_p$ contains the indices of sequences assigned to family $p$. At each step, we sample $P$ families and then sample $K$ sequences from each selected bucket, forming a batch of size $PK$. This yields multiple positives per anchor while preserving diverse negatives across families.

\subsection{Multi-Positive Contrastive Objective}
\label{sec:method_loss}
For each sequence $s_i$ in a batch $\mathcal{B}$, we compute the pooled representation from the adapted encoder:
\begin{equation}
\mathbf{z}_i=\mathbf{z}_{s_i}=\mathrm{Pool}(f_{\theta+\Delta\theta}(s_i)).
\end{equation}
We then map it through a projection head $g_{\phi}$ and apply $\ell_2$ normalization:
\begin{equation}
\mathbf{h}_i=\frac{g_{\phi}(\mathbf{z}_i)}{\|g_{\phi}(\mathbf{z}_i)\|_2},
\end{equation}
where $\|\cdot\|_2$ denotes the $\ell_2$ norm. Pairwise similarity is then computed by temperature-scaled cosine similarity with temperature $\tau$:
\begin{equation}
\mathrm{sim}_{ij}=\frac{\mathbf{h}_i^\top \mathbf{h}_j}{\tau}.
\end{equation}

SAVR optimizes a multi-positive InfoNCE objective, where each anchor is contrasted against all other sequences in the batch and positives are defined by $\mathcal{P}_{\mathcal{B}}(i)$:
\begin{equation}
\mathcal{L}
=
-\frac{1}{|\mathcal{I}|}
\sum_{i\in \mathcal{I}}
\frac{1}{|\mathcal{P}_{\mathcal{B}}(i)|}
\sum_{j\in \mathcal{P}_{\mathcal{B}}(i)}
\log
\frac{\exp(\mathrm{sim}_{ij})}
{\sum\limits_{k\in \mathcal{B}\setminus\{i\}} \exp(\mathrm{sim}_{ik})},
\label{eq:savr_loss}
\end{equation}
where $\mathcal{I}=\{i\in\mathcal{B}:\ |\mathcal{P}_{\mathcal{B}}(i)|>0\}$ denotes anchors with at least one in-batch positive.

The projection head $g_{\phi}$ is used only for contrastive post-training and discarded afterward. At inference time, SAVR uses the adapted encoder's pooled backbone embedding $\mathbf{z}_s$ as the final sequence representation. Algorithm~\ref{alg:savr} summarizes SAVR post-training. 

\begin{algorithm}[tb]
\renewcommand{\arraystretch}{1.15}
\caption{SAVR: Pfam-guided contrastive post-training with LoRA}
\label{alg:savr}
\begin{algorithmic}[1]
\STATE {\bfseries Input:} dataset $\mathcal{D}=\{(s_i,F(s_i))\}_{i=1}^{N}$; sampler $(P,K)$; temperature $\tau$
\STATE {\bfseries Model:} frozen encoder $f_{\theta}$ with LoRA adapters $\Delta\theta$; projection head $g_{\phi}$
\STATE Construct Pfam buckets $\{\mathcal{B}_p\}$ where $\mathcal{B}_p=\{i:\ p\in F(s_i)\}$
\FOR{each training step}
    \STATE Sample $P$ families $\{p_m\}_{m=1}^{P}$
    \STATE Form batch $\mathcal{B}\leftarrow \bigcup_{m=1}^{P} \textsc{SampleK}(\mathcal{B}_{p_m},K)$ \hfill ($|\mathcal{B}|=PK$)
    \STATE Compute pooled embeddings $\mathbf{z}_i \leftarrow \mathrm{Pool}(f_{\theta+\Delta\theta}(s_i))$, $\forall i\in\mathcal{B}$
    \STATE Project and normalize $\mathbf{h}_i \leftarrow g_{\phi}(\mathbf{z}_i)/\|g_{\phi}(\mathbf{z}_i)\|_2$, $\forall i\in\mathcal{B}$
    \STATE (DDP) All-gather $\{\mathbf{h}_i\}$ across devices to obtain global batch representations
    \STATE For each anchor $i$, set $\mathcal{P}_{\mathcal{B}}(i)\leftarrow \{j\in\mathcal{B}\setminus\{i\}:\ F(s_i)\cap F(s_j)\neq \emptyset\}$
    \STATE Compute loss $\mathcal{L}$ using Eq.~\eqref{eq:savr_loss} over anchors with $|\mathcal{P}_{\mathcal{B}}(i)|>0$
    \STATE Update $(\Delta\theta,\phi)$ by gradient descent on $\mathcal{L}$
\ENDFOR
\STATE {\bfseries Output:} adapted encoder parameters $(\theta+\Delta\theta)$
\end{algorithmic}
\end{algorithm}
\vspace{-0.6em}

\section{Experimental Validation}
\label{sec:exp}
We evaluate SAVR by testing whether it rectifies representation geometry (Section~\ref{sec:geom}), improves downstream function prediction (Section~\ref{sec:downstream}), and outperforms key controls for generic contrastive learning and geometry-only post-processing (Section~\ref{sec:exp:ablation}).

\subsection{Experimental Setup}
\label{subsec:experiment_setup}
\paragraph{Post-training dataset and optimization}
We construct a family-based contrastive dataset from UniRef50~\cite{uniprot2021uniprot} sequences annotated with Pfam domains using InterProScan~\cite{blum2021interpro,interproscan}. We retain Pfam families with at least $70$ annotated sequences and randomly sample $70$ sequences from each retained family, yielding approximately $16,000$ families with fixed per-family support for a $50/10/10$ train/validation/test split. During post-training, positive pairs are defined by shared family annotations, while other in-batch pairs are treated as negatives. During post-training, Pfam annotations are used only to define pair membership; the model never predicts Pfam identifiers, and no EC/GO labels or task-specific annotations are used (Refer to Appendix~\ref{app:sec:leakage} for more detailed discussion). We optimize Eq.~\eqref{eq:savr_loss} with AdamW and use distributed all-gather over $\{\mathbf{h}_i\}$ to form a global contrastive batch with more negatives. Additional model and hyperparameter details are provided in Appendix~\ref{appendix:implementation_details}.

\paragraph{Evaluation Protocol} 
We evaluate SAVR on EC and GO benchmarks under frozen-backbone settings to isolate representation quality. For the main results, we use ESM2-650M~\cite{lin2023evolutionary} as the backbone and train only a lightweight one-hidden-layer MLP on top of frozen pooled representations (see Appendix~\ref{sec:app:mlp}). For EC prediction, we adopt the open benchmark from \citet{yu2023enzyme} and follow the training setting used by  \citet{yang2025interpretable}. For GO prediction, we follow the benchmark configurations and training procedures from \citet{wang2025dpfunc} and \citet{yang2025dpgok}. Implementations are conducted with VenusFactory~\cite{tan2025venusfactory}, which provides the required benchmark resources, model components, training utilities, and evaluation pipeline. More details are provided in Appendix~\ref{appendix:exp_implementation}.


\subsection{Representation Geometry Analysis}
\label{sec:geom}
We compute geometry metrics on embeddings from $16$k sequences uniformly sampled from the Pfam test split. We analyze the covariance spectrum of pooled embeddings after row-wise $\ell_2$ normalization, which controls for per-sequence embedding norm and isolates directional structure. See Appendices~\ref{appendix:normalization} and~\ref{appendix:raw} for details on the normalization procedure and raw-embedding results.



\begin{figure}[t]
    \centering

    \begin{subfigure}[t]{0.54\linewidth}
        \centering
        \includegraphics[width=\linewidth]{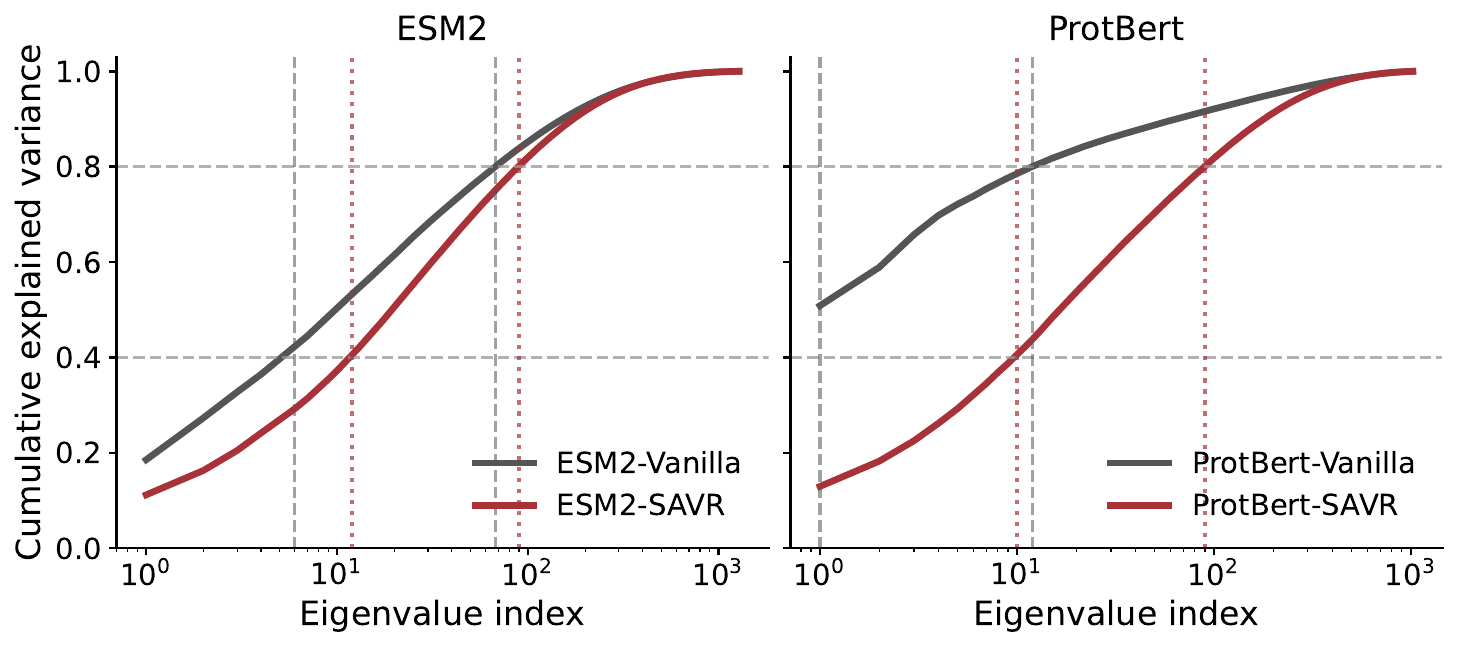}
        \label{fig:cumvar}
    \end{subfigure}
    \hfill
    \begin{subfigure}[t]{0.44\linewidth}
        \centering
        \includegraphics[width=\linewidth]{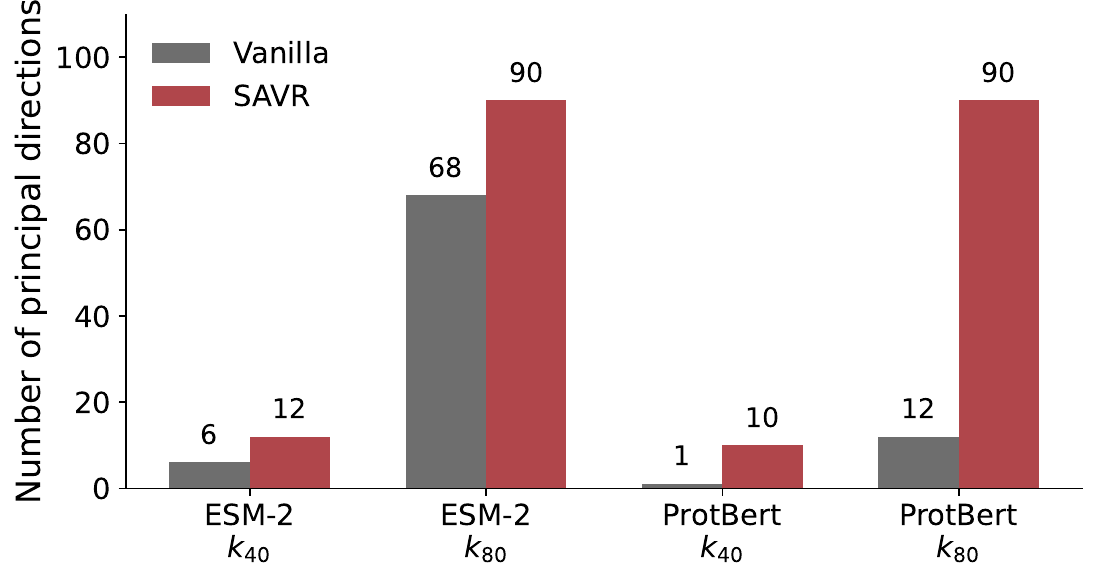}
        \label{fig:k_boundary}
    \end{subfigure}

    \caption{Spectral redistribution induced by SAVR. Left: SAVR spreads cumulative explained variance over more eigenvalue directions. Dotted vertical lines mark the eigenvalue indices at which each representation reaches the 40\% and 80\% cumulative variance thresholds, with gray for Vanilla and red for SAVR. Right: the number of principal directions required to explain 40\% and 80\% variance increases after SAVR, indicating reduced dominance by the leading eigendirections.} 
    \label{fig:spectral_redistribution}
\end{figure}

\paragraph{SAVR induces consistent geometric rectification across backbones.}
Figure~\ref{fig:spectral_redistribution} shows a consistent pattern across ESM-2 and ProtBert. After SAVR, the cumulative explained-variance curves shift to the right and the number of principal directions required to explain fixed fractions of total variance increases. This indicates a less concentrated covariance spectrum, where pooled sequence embeddings are no longer dominated by a few leading eigendirections. Such geometry is more favorable for lightweight downstream readouts, as protein-level functional variation is less likely to be compressed into hard-to-access low-variance directions.

\paragraph{SAVR spreads variance beyond the leading eigendirections.}
\begin{wraptable}{r}{0.58\textwidth}
\centering
\vspace{-1.5em}
\small
\setlength{\tabcolsep}{4pt}
\renewcommand{\arraystretch}{1.1}
\caption{Representation geometry diagnostics on row-wise $\ell_2$ normalized pooled embeddings. RankMe measures effective dimensionality, Top-10 reports variance concentration in the leading directions, and piecewise spectral exponents ($\alpha_h,\alpha_m,\alpha_t$) report local spectrum changes.}
\label{tab:geometry-core-l2}
\resizebox{\linewidth}{!}{%
\begin{tabular}{l@{\hspace{3pt}}c@{\hspace{4pt}}cccc}
\toprule
Method & RankMe$\uparrow$ & Top-10$\downarrow$ & $\alpha_{h}$ & $\alpha_{m}$ & $\alpha_{t}$\\
\midrule
ESM-2 & 77.9 & 0.503 & 1.110 & 1.108 & 2.434 \\
ESM-SAVR & \textbf{122.5} & \textbf{0.372} & 0.749 & 1.110 & 2.775 \\
\midrule
ProtBert & 15.8 & 0.785 & -- & 1.474 & 1.556 \\
PB-SAVR & \textbf{113.2} & \textbf{0.406} & 0.788 & 1.115 & 2.510 \\
\bottomrule
\end{tabular}
}
\vspace{-1.5em}
\end{wraptable}
The variance-threshold boundaries further localize this effect. The $40\%$ boundary moves from $6$ to $12$ for ESM2-650M and from $1$ to $10$ for ProtBert. The $80\%$ boundary shows an even larger shift, moving from $68$ to $90$ for ESM2-650M and from $12$ to $90$ for ProtBert. These shifts show that SAVR reduces head-spectrum dominance rather than merely changing the overall embedding scale. Table~\ref{tab:geometry-core-l2} provide a consistent view: RankMe increases, the Top-10 variance ratio decreases, and $\alpha_h$ becomes smaller after SAVR, indicating a flatter leading spectrum. Meanwhile, $\alpha_t$ increases mainly in the residual low-variance tail, suggesting that SAVR does not uniformly flatten the entire spectrum.




\subsection{Downstream Protein-Level Function Prediction}
\label{sec:downstream}
We next evaluate SAVR-rectified representations on downstream protein-level function prediction, including $5$ popular open benchmarks on Enzyme Commission (EC) and Gene Ontology (GO). 

\subsubsection{Enzyme Commission Prediction}
We evaluate SAVR-refined ESM2-650M on EC number prediction following the CLEAN~\cite{yu2023enzyme} benchmark protocol. Classifiers are trained on the CLEAN \texttt{split100} training set and evaluated on \textbf{New-392} and \textbf{Price-149}, which contain $392$ newly identified and $149$ experimentally verified proteins, respectively. We use two one-hidden-layer MLP heads that differ only in pooling: mean pooling or a learned attention-pooling module implemented by a one-layer MLP scorer followed by softmax over tokens. Following CLEAN, we report weighted precision, recall, and F1, and compare against prior state-of-the-art EC prediction methods.

\textbf{SAVR achieves the strongest EC performance with improved recall.}
Table~\ref{tab:ec} shows that SAVR achieves the strongest overall EC performance on both \textbf{New-392} and \textbf{Price-149}. Notably, SAVR exhibits substantially higher recall than prior methods, a desirable property for biological screening tasks with large negative search spaces, suggesting that the rectified representations expose function-relevant signals more effectively to lightweight classifiers. Mean and attention pooling mainly control the precision-recall trade-off: mean pooling favors recall, whereas attention pooling improves precision. Since both variants remain competitive, the main improvement comes from the SAVR representation rather than the pooling choice.

\begin{table*}[t]
\centering
\caption{EC prediction performance on \textbf{New-392} and \textbf{Price-149}.}
\label{tab:ec}
\resizebox{\textwidth}{!}{%
\begin{tabular}{l ccc ccc}
\toprule
&
\multicolumn{3}{c}{\textbf{New-392}} &
\multicolumn{3}{c}{\textbf{Price-149}} \\
\cmidrule(lr){2-4}\cmidrule(lr){5-7}
\textbf{Method} & Precision & Recall & F1 Score
& Precision & Recall & F1 Score \\
\midrule
\textsc{DEEPre}~\cite{li2018deepre}      & --    & --    & --    & 0.042 & 0.040 & 0.039 \\
\textsc{BLASTp}~\cite{altschul1990basic}      & --    & --    & --    & 0.508 & 0.375 & 0.385 \\
\textsc{ECPred} \cite{dalkiran2018ecpred} & 0.178 & 0.095 & 0.100 & 0.020 & 0.020 & 0.020 \\
\textsc{DeepEC} \cite{ryu2019deep} & 0.298 & 0.217 & 0.230 & 0.118 & 0.072 & 0.085 \\
\textsc{ProtInfer} \cite{sanderson2023proteinfer} & 0.409 & 0.284 & 0.309 & 0.243 & 0.138 & 0.166 \\
\textsc{DeepECTransformer} \cite{kim2023functional} & 0.427 & 0.326 & 0.335 & 0.526 & 0.303 & 0.351 \\
\textsc{CLEAN} \cite{yu2023enzyme} & \underline{0.597} & 0.481 & 0.499 & \underline{0.584} & 0.467 & 0.495 \\
\textsc{ProtDETR} \cite{yang2025interpretable}    & 0.594 & 0.608 & 0.577 & 0.577 & 0.507 & 0.508 \\
\midrule
\textsc{SAVR}        & 0.589 & \textbf{0.829} & \textbf{0.616} & 0.574 & \textbf{0.678} & \underline{0.587} \\
\textsc{SAVR (Attn)} & \textbf{0.632} & \underline{0.746} & \underline{0.605} & \textbf{0.647} & \underline{0.625} & \textbf{0.607} \\
\bottomrule\\[-2.5mm]
\multicolumn{7}{l}{$^\dagger$\footnotesize{The \textbf{best} and \underline{second-best} performed results are highlighted}.}\\
\multicolumn{7}{l}{$^\star$\footnotesize{ECPRED, DEEPEC, PROTINFER and CLEAN results are from CLEAN~\cite{yu2023enzyme}, other results from PROTDETR~\cite{yang2025interpretable}}.}
\end{tabular}
}
\end{table*}

\subsubsection{Gene Ontology Prediction}
\begin{table*}[t]
\centering
\caption{Performance comparison on the three \textbf{GO benchmark}.}
\label{tab:go-benchmark}
\resizebox{0.95\textwidth}{!}{%
\begin{tabular}{l l cc cc cc}
\toprule
&
&
\multicolumn{2}{c}{\textbf{GO-MF}} &
\multicolumn{2}{c}{\textbf{GO-BP}} &
\multicolumn{2}{c}{\textbf{GO-CC}} \\
\cmidrule(lr){3-4}\cmidrule(lr){5-6}\cmidrule(lr){7-8}
\textbf{Method} & \textbf{In ference Input}
& F$_\text{max}$ & AUPR
& F$_\text{max}$ & AUPR
& F$_\text{max}$ & AUPR \\
\midrule
\textsc{TALE} \cite{cao2021tale}   & single sequence
& 0.446 & 0.365 & 0.270 & 0.199 & 0.547 & 0.534 \\
\textsc{ATGO} \cite{zhu2022integrating}   & single sequence
& 0.496 & 0.500 & 0.374 & 0.323 & 0.630 & 0.646 \\
\textsc{BlastKNN} \cite{madden2013blast}  & retrieval
& 0.591 & 0.495 & 0.411 & 0.290 & 0.580 & 0.451 \\
\textsc{DeepGoZero} \cite{kulmanov2022deepgozero}   & GO prior
& 0.608 & 0.552 & 0.397 & 0.329 & 0.616 & 0.597 \\
\textsc{DeepGoSE} \cite{kulmanov2023deepgo}  & GO prior
& 0.588 & 0.578 & 0.422 & 0.371 & 0.648 & 0.633 \\
\textsc{DPGOK}~\cite{yang2025dpgok}  & GO + domain prior
& \textcolor{violet}{\textbf{0.620}} & \textcolor{violet}{\textbf{0.633}}
& \textbf{0.438} & \textcolor{violet}{\textbf{0.383}}
& \textcolor{violet}{\textbf{0.680}} & \textcolor{red}{\textbf{0.711}} \\
\textsc{TALE+}~\cite{cao2021tale}  & ensemble
& 0.572 & 0.512 & 0.411 & 0.336 & 0.613 & 0.607 \\
\textsc{ATGO+}~\cite{zhu2022integrating}  & ensemble
& 0.610 & \textbf{0.608}
& \textcolor{violet}{\textbf{0.444}} & \textbf{0.380}
& 0.645 & 0.672 \\
\textsc{DPGOK+}~\cite{yang2025dpgok} & ensemble
& \textcolor{red}{\textbf{0.629}} & \textcolor{red}{\textbf{0.635}}
& \textcolor{red}{\textbf{0.458}} & \textcolor{red}{\textbf{0.400}}
& \textcolor{red}{\textbf{0.681}} & \textcolor{violet}{\textbf{0.706}} \\
\midrule
\textsc{SAVR}   & single sequence
& \textbf{0.613} & 0.605
& 0.421 & 0.356
& \textbf{0.661} & \textbf{0.690} \\
\bottomrule\\[-2.5mm]
\multicolumn{8}{l}{$^\dagger$\footnotesize{The top three results are highlighted as \textcolor{red}{\textbf{First}}, \textcolor{violet}{\textbf{Second}}, and \textbf{Third}.}}
\end{tabular}
}
\end{table*}

We further evaluate SAVR-trained ESM2-650M on GO prediction across Molecular Function (MF), Biological Process (BP), and Cellular Component (CC). The benchmark is curated from UniProtKB/Swiss-Prot snapshots and contains only proteins with experimentally supported GO annotations. It uses a time-based split: proteins released before Jan 2022 are used for training, proteins released between Jan 2022 and Jan 2023 for validation, and proteins released after Jan 2023 for testing. Consistent with DPGOK \cite{yang2025dpgok}, we train and evaluate each ontology branch independently, freeze the SAVR backbone, train a one-hidden-layer MLP classifier, and report F$_\text{max}$ and AUPR.

\textbf{SAVR outperforms single-sequence baselines and approaches stronger GO-specific systems.}
Table~\ref{tab:go-benchmark} shows that SAVR achieves competitive GO performance under a constrained single-sequence inference setting. Unlike retrieval-based methods, ontology-aware methods, and ensemble variants, SAVR predicts directly from the query-sequence embedding, without annotated-reference retrieval, GO hierarchy or label-structure priors, auxiliary domain knowledge, or model aggregation. Under comparable single-sequence inference, SAVR consistently outperforms TALE and ATGO across all ontology branches. Compared with stronger hybrid or ensemble systems, SAVR remains close on most metrics and reaches top-three performance on \textbf{GO-MF} and \textbf{GO-CC}. This indicates that representation rectification provides a strong parametric readout signal for GO prediction, even without extra inference-time resources.

\subsection{Controlled Analyses}
\label{sec:exp:ablation}
We conduct controlled analyses to isolate the source of SAVR's improvements. We first vary the positive-pair construction rule under the same contrastive post-training objective, then compare SAVR with downstream adaptation baselines, and finally test whether post-hoc geometry normalization alone is sufficient for explaining the gains.


\subsubsection{Positive-Pair Construction Control}
We compare SAVR with two alternative contrastive post-training strategies under the same backbone, objective, and optimization setting. \textbf{Random CL} uses randomly assigned positive pairs, testing whether contrastive learning alone is sufficient. \textbf{SI $\geq 30\%$ CL} defines positives by MMseqs2 sequence identity above $30\%$, testing a fixed-threshold homology-pairing rule. SAVR instead uses family co-membership to define homology-guided positives.

\begin{wraptable}{r}{0.53\textwidth}
\centering
\vspace{-1.0em}
\small
\setlength{\tabcolsep}{4pt}
\renewcommand{\arraystretch}{1.12}
\caption{Comparison of different contrastive post-training strategies.}
\label{tab:contrastive_ablation}
\resizebox{\linewidth}{!}{%
\begin{tabular}{lccc}
\toprule
Method & RankMe$\uparrow$ & Top-10$\downarrow$ & $\alpha_h$ \\
\midrule
Random CL        & 10.9 (collapse) & 0.865 & --     \\
SI $\geq$30\% CL & 94.0            & 0.446 & 1.080 \\
SAVR             & \textbf{122.5}  & \textbf{0.372} & \textbf{0.749} \\
\bottomrule
\end{tabular}
}
\vspace{-0.3em}
\end{wraptable}

\textbf{Co-membership-guided pairing is necessary for useful rectification.}
Table~\ref{tab:contrastive_ablation} shows that random positives collapse the representation geometry, indicating that the gains are not caused by contrastive learning alone. Fixed sequence-identity positives improve the geometry, confirming that homology-guided post-training is useful. SAVR performs best across RankMe, Top-10, and $\alpha_h$, suggesting that family-aware pairing provides a stronger and less threshold-sensitive grouping rule than a manually selected sequence-identity cutoff. The corresponding EC results, reported in Appendix~\ref{appendix:contrastive_ablation}, follow the same trend.


\subsubsection{Fine-Tuning Control}

To disentangle the effect of SAVR post-training from task-specific adaptation, we assess four fine-tuning strategies, including: \textbf{Freeze} trains only a two-layer MLP classifier on frozen backbone embeddings. \textbf{LoRA} performs task-specific parameter-efficient low-rank adaptation during downstream training. \textbf{SES-Adpt}~\cite{tan2024simple} is a structure-aware adapter baseline using both sequence and structural inputs. \textbf{SAVR} uses the frozen SAVR post-trained backbone with the same classifier as \textbf{Freeze}. All settings use the default training and evaluation pipeline from VenusFactory~\cite{tan2025venusfactory}.

\begin{wraptable}{r}{0.6\textwidth}
\centering
\vspace{-1.5em}
\small
\setlength{\tabcolsep}{4pt}
\renewcommand{\arraystretch}{1.15}
\caption{Performance under different fine-tuning strategies.}
\label{tab:venus_results}
\resizebox{\linewidth}{!}{%
\begin{tabular}{l@{\hspace{4pt}}lcccc}
\toprule
Model & FT & EC & GO-BP & GO-CC & GO-MF \\
\midrule
\multirow{4}{*}{ESM2}
  & Freeze      & 0.843 & \underline{0.484} & \textbf{0.577} & 0.640 \\
  & LoRA        & \underline{0.852} & 0.483 & 0.461 & \underline{0.664} \\
  & SES-Adpt    & 0.848 & 0.466 & 0.526 & 0.634 \\
  & SAVR        & \textbf{0.865} & \textbf{0.518} & \underline{0.568} & \textbf{0.687} \\
\midrule
\multirow{4}{*}{ProtBert}
  & Freeze      & 0.701 & \underline{0.420} & \underline{0.546} & 0.523 \\
  & LoRA        & 0.764 & 0.245 & 0.316 & 0.161 \\
  & SES-Adpt    & \underline{0.766} & 0.416 & 0.495 & \underline{0.546} \\
  & SAVR        & \textbf{0.843} & \textbf{0.490} & \textbf{0.553} & \textbf{0.657} \\
\bottomrule
\end{tabular}
}
\vspace{-1em}
\end{wraptable}

\paragraph{SAVR improves representation quality beyond task-time adaptation.}
Table~\ref{tab:venus_results} shows that SAVR achieves the strongest and most consistent performance across EC and GO with both backbones. The gain is especially pronounced for ProtBert, where SAVR yields a large improvement and substantially narrows the gap between the pre-trained ProtBert performance to the stronger ESM2-650M backbone. In contrast, task-time LoRA is unstable across tasks and can severely degrade ProtBert GO performance, suggesting that downstream adaptation with limited task labels is not a reliable substitute for representation-level post-training. These results indicate that SAVR and downstream adaptation operate at different levels: SAVR improves the backbone representation before task labels are seen, while LoRA or adapters specialize the model to a specific endpoint.



\subsubsection{Geometry-Only Control}


\begin{wrapfigure}{r}{0.55\textwidth}
  \vspace{-1.5em}
  \centering
  \includegraphics[width=\linewidth]{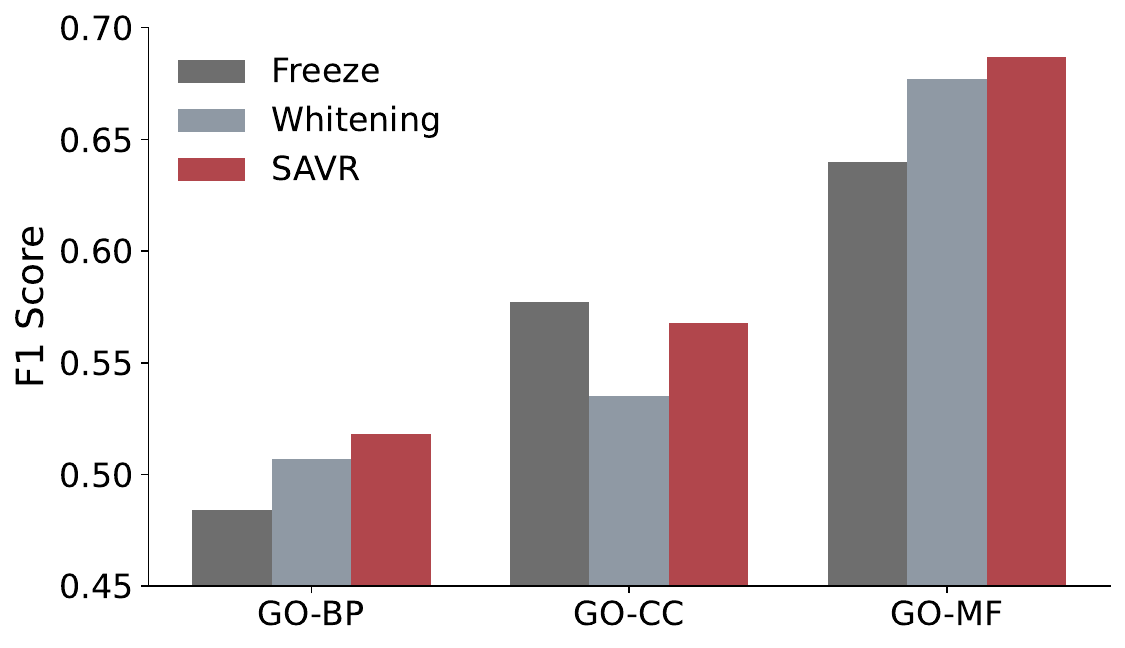}
  \caption{GO branch-wise F1 scores comparing the frozen baseline, post-hoc whitening, and SAVR.}
  \vspace{-2.em}
  \label{fig:whiten_ablation}
\end{wrapfigure}

Furthermore, we test whether downstream gains can be explained by geometry normalization alone. As a post-hoc control, we apply PCA whitening to vanilla ESM2 embeddings by fitting the transform on training-set embeddings and applying the same mapping to both training and test embeddings. We then train the same MLP classifier with identical optimization settings on the VenusFactory GO benchmark.

\textbf{SAVR is not reducible to global covariance re-scaling.}
Figure~\ref{fig:whiten_ablation} shows that whitening provides only partial and inconsistent improvements across GO branches. Although whitening can reduce second-order anisotropy, it does not match SAVR's performance, indicating that the gains require more than global covariance normalization. This supports the view that SAVR improves downstream readout through informative post-training, not merely by re-scaling the embedding spectrum.

\section{Conclusion and Limitations}

Pretrained PLMs provide reusable sequence encoders for downstream biological prediction tasks, but token-level likelihood objectives can produce embeddings dominated by a few principal directions, making protein-level functional variation difficult to access with lightweight readouts. We introduced SAVR, a parameter-efficient contrastive post-training framework that reshapes PLM representations using homology-guided sequence pairs without explicitly using EC/GO labels. Across backbones, SAVR reduces variance concentration, increases effective dimensionality, and improves protein function prediction, achieving strong EC performance and competitive GO results under single-sequence inference. Controlled analyses further show that these gains are not explained by contrastive learning alone, fixed sequence-identity pairing, or post-hoc covariance re-scaling.

\textbf{Limitations.} The main limitation of SAVR is that Pfam co-membership is a coarse protein-level signal. It does not explicitly model domain order, domain architecture, or residue-level functional sites, and our evaluation is scoped to protein-level function prediction rather than universal improvement across all protein tasks. Future work could extend SAVR with finer structural or residue-level signals.

\bibliographystyle{unsrtnat}
\bibliography{ref}

@article{zhou2024mlife,
  title={Protein engineering in the deep learning era},
  author={Zhou, Bingxin and Tan, Yang and Hu, Yutong and Zheng, Lirong and Zhong, Bozitao and Hong, Liang},
  journal={mLife},
  volume={3},
  number={4},
  pages={477--491},
  year={2024},
  publisher={Wiley Online Library}
}

@article{interproscan,
    author = {Jones, Philip and Binns, David and Chang, Hsin-Yu and Fraser, Matthew and Li, Weizhong and McAnulla, Craig and McWilliam, Hamish and Maslen, John and Mitchell, Alex and Nuka, Gift and Pesseat, Sebastien and Quinn, Antony F. and Sangrador-Vegas, Amaia and Scheremetjew, Maxim and Yong, Siew-Yit and Lopez, Rodrigo and Hunter, Sarah},
    title = {InterProScan 5: genome-scale protein function classification},
    journal = {Bioinformatics},
    volume = {30},
    number = {9},
    pages = {1236-1240},
    year = {2014},
    month = {01},
    issn = {1367-4803},
    doi = {10.1093/bioinformatics/btu031},
    url = {https://doi.org/10.1093/bioinformatics/btu031},
    eprint = {https://academic.oup.com/bioinformatics/article-pdf/30/9/1236/48923139/bioinformatics_30_9_1236.pdf},
}

@article{blum2021interpro,
  title={The InterPro protein families and domains database: 20 years on},
  author={Blum, Matthias and Chang, Hsin-Yu and Chuguransky, Sara and Grego, Tiago and Kandasaamy, Swaathi and Mitchell, Alex and Nuka, Gift and Paysan-Lafosse, Typhaine and Qureshi, Matloob and Raj, Shriya and others},
  journal={Nucleic acids research},
  volume={49},
  number={D1},
  pages={D344--D354},
  year={2021},
  publisher={Oxford University Press}
}

@article{lin2023evolutionary,
  title={Evolutionary-scale prediction of atomic-level protein structure with a language model},
  author={Lin, Zeming and Akin, Halil and Rao, Roshan and Hie, Brian and Zhu, Zhongkai and Lu, Wenting and Smetanin, Nikita and Verkuil, Robert and Kabeli, Ori and Shmueli, Yaniv and others},
  journal={Science},
  volume={379},
  number={6637},
  pages={1123--1130},
  year={2023},
  publisher={American Association for the Advancement of Science}
}

@article{elnaggar2021prottrans,
  title={ProtTrans: towards cracking the language of life’s code through self-supervised learning},
  author={Elnaggar, Ahmed and Heinzinger, Michael and Dallago, Christian and Rehawi, Ghalia and Wang, Yu and Jones, Llion and Gibbs, Tom and Feher, Tamas and Angerer, Christoph and Steinegger, Martin and others},
  journal={IEEE Transactions on Pattern Analysis and Machine Intelligence},
  volume={44},
  pages={7112--7127},
  year={2021}
}

@article{radivojac2013large,
  title={A large-scale evaluation of computational protein function prediction},
  author={Radivojac, Predrag and Clark, Wyatt T and Oron, Tal Ronnen and Schnoes, Alexandra M and Wittkop, Tobias and Sokolov, Artem and Graim, Kiley and Funk, Christopher and Verspoor, Karin and Ben-Hur, Asa and others},
  journal={Nature methods},
  volume={10},
  number={3},
  pages={221--227},
  year={2013},
  publisher={Nature Publishing Group US New York}
}

@article{jiang2016expanded,
  title={An expanded evaluation of protein function prediction methods shows an improvement in accuracy},
  author={Jiang, Yuxiang and Oron, Tal Ronnen and Clark, Wyatt T and Bankapur, Asma R and D’Andrea, Daniel and Lepore, Rosalba and Funk, Christopher S and Kahanda, Indika and Verspoor, Karin M and Ben-Hur, Asa and others},
  journal={Genome biology},
  volume={17},
  number={1},
  pages={184},
  year={2016},
  publisher={Springer}
}

@article{zhou2019cafa,
  title={The CAFA challenge reports improved protein function prediction and new functional annotations for hundreds of genes through experimental screens},
  author={Zhou, Naihui and Jiang, Yuxiang and Bergquist, Timothy R and Lee, Alexandra J and Kacsoh, Balint Z and Crocker, Alex W and Lewis, Kimberley A and Georghiou, George and Nguyen, Huy N and Hamid, Md Nafiz and others},
  journal={Genome biology},
  volume={20},
  number={1},
  pages={244},
  year={2019},
  publisher={Springer}
}

@article{jumper2021highly,
  title={Highly accurate protein structure prediction with AlphaFold},
  author={Jumper, John and Evans, Richard and Pritzel, Alexander and Green, Tim and Figurnov, Michael and Ronneberger, Olaf and Tunyasuvunakool, Kathryn and Bates, Russ and {\v{Z}}{\'\i}dek, Augustin and Potapenko, Anna and others},
  journal={nature},
  volume={596},
  number={7873},
  pages={583--589},
  year={2021},
  publisher={Nature Publishing Group UK London}
}

@article{baek2021accurate,
  title={Accurate prediction of protein structures and interactions using a three-track neural network},
  author={Baek, Minkyung and DiMaio, Frank and Anishchenko, Ivan and Dauparas, Justas and Ovchinnikov, Sergey and Lee, Gyu Rie and Wang, Jue and Cong, Qian and Kinch, Lisa N and Schaeffer, R Dustin and others},
  journal={Science},
  volume={373},
  number={6557},
  pages={871--876},
  year={2021},
  publisher={American Association for the Advancement of Science}
}

@article{yu2023enzyme,
  title={Enzyme function prediction using contrastive learning},
  author={Yu, Tianhao and Cui, Haiyang and Li, Jianan Canal and Luo, Yunan and Jiang, Guangde and Zhao, Huimin},
  journal={Science},
  volume={379},
  number={6639},
  pages={1358--1363},
  year={2023},
  publisher={American Association for the Advancement of Science}
}

@article{li2025tracing,
  title={Tracing the representation geometry of language models from pretraining to post-training},
  author={Li, Melody Zixuan and Agrawal, Kumar Krishna and Ghosh, Arna and Teru, Komal Kumar and Santoro, Adam and Lajoie, Guillaume and Richards, Blake A},
  journal={arXiv preprint arXiv:2509.23024},
  year={2025}
}

@misc{ethayarajh2019contextual,
      title={How Contextual are Contextualized Word Representations? Comparing the Geometry of BERT, ELMo, and GPT-2 Embeddings}, 
      author={Kawin Ethayarajh},
      year={2019},
      eprint={1909.00512},
      archivePrefix={arXiv},
      primaryClass={cs.CL},
      url={https://arxiv.org/abs/1909.00512}, 
}

@inproceedings{garrido2023rankme,
  title={Rankme: Assessing the downstream performance of pretrained self-supervised representations by their rank},
  author={Garrido, Quentin and Balestriero, Randall and Najman, Laurent and Lecun, Yann},
  booktitle={International conference on machine learning},
  pages={10929--10974},
  year={2023},
  organization={PMLR}
}

@article{huang2021whiteningbert,
  title={WhiteningBERT: An easy unsupervised sentence embedding approach},
  author={Huang, Junjie and Tang, Duyu and Zhong, Wanjun and Lu, Shuai and Shou, Linjun and Gong, Ming and Jiang, Daxin and Duan, Nan},
  journal={arXiv preprint arXiv:2104.01767},
  year={2021}
}

@article{yang2025interpretable,
  title={Interpretable Enzyme Function Prediction via Residue-Level Detection},
  author={Yang, Zhao and Su, Bing and Chen, Jiahao and Wen, Ji-Rong},
  journal={arXiv preprint arXiv:2501.05644},
  year={2025}
}

@article{agrawal2022alreq,
  title={$\alpha$-ReQ: Assessing Representation Quality in Self-Supervised Learning by measuring eigenspectrum decay},
  author={Agrawal, Kumar K and Mondal, Arnab Kumar and Ghosh, Arna and Richards, Blake},
  journal={Advances in Neural Information Processing Systems},
  volume={35},
  pages={17626--17638},
  year={2022}
}

@article{zucchet2025language,
  title={How do language models learn facts? Dynamics, curricula and hallucinations},
  author={Zucchet, Nicolas and Bornschein, J{\"o}rg and Chan, Stephanie and Lampinen, Andrew and Pascanu, Razvan and De, Soham},
  journal={arXiv preprint arXiv:2503.21676},
  year={2025}
}

@article{power2022grokking,
  title={Grokking: Generalization beyond overfitting on small algorithmic datasets},
  author={Power, Alethea and Burda, Yuri and Edwards, Harri and Babuschkin, Igor and Misra, Vedant},
  journal={arXiv preprint arXiv:2201.02177},
  year={2022}
}

@inproceedings{refinetti2023neural,
  title={Neural networks trained with sgd learn distributions of increasing complexity},
  author={Refinetti, Maria and Ingrosso, Alessandro and Goldt, Sebastian},
  booktitle={International Conference on Machine Learning},
  pages={28843--28863},
  year={2023},
  organization={PMLR}
}

@article{ouyang2022training,
  title={Training language models to follow instructions with human feedback},
  author={Ouyang, Long and Wu, Jeffrey and Jiang, Xu and Almeida, Diogo and Wainwright, Carroll and Mishkin, Pamela and Zhang, Chong and Agarwal, Sandhini and Slama, Katarina and Ray, Alex and others},
  journal={Advances in neural information processing systems},
  volume={35},
  pages={27730--27744},
  year={2022}
}

@inproceedings{houlsby2019parameter,
  title={Parameter-efficient transfer learning for NLP},
  author={Houlsby, Neil and Giurgiu, Andrei and Jastrzebski, Stanislaw and Morrone, Bruna and De Laroussilhe, Quentin and Gesmundo, Andrea and Attariyan, Mona and Gelly, Sylvain},
  booktitle={International conference on machine learning},
  pages={2790--2799},
  year={2019},
  organization={PMLR}
}

@article{rafailov2023direct,
  title={Direct preference optimization: Your language model is secretly a reward model},
  author={Rafailov, Rafael and Sharma, Archit and Mitchell, Eric and Manning, Christopher D and Ermon, Stefano and Finn, Chelsea},
  journal={Advances in neural information processing systems},
  volume={36},
  pages={53728--53741},
  year={2023}
}

@article{shao2024deepseekmath,
  title={Deepseekmath: Pushing the limits of mathematical reasoning in open language models},
  author={Shao, Zhihong and Wang, Peiyi and Zhu, Qihao and Xu, Runxin and Song, Junxiao and Bi, Xiao and Zhang, Haowei and Zhang, Mingchuan and Li, YK and Wu, Yang and others},
  journal={arXiv preprint arXiv:2402.03300},
  year={2024}
}

@article{hu2022lora,
  title={Lora: Low-rank adaptation of large language models.},
  author={Hu, Edward J and Shen, Yelong and Wallis, Phillip and Allen-Zhu, Zeyuan and Li, Yuanzhi and Wang, Shean and Wang, Lu and Chen, Weizhu and others},
  journal={ICLR},
  volume={1},
  number={2},
  pages={3},
  year={2022}
}

@inproceedings{chen2020simple,
  title={A simple framework for contrastive learning of visual representations},
  author={Chen, Ting and Kornblith, Simon and Norouzi, Mohammad and Hinton, Geoffrey},
  booktitle={International conference on machine learning},
  pages={1597--1607},
  year={2020},
  organization={PmLR}
}

@article{khosla2020supervised,
  title={Supervised contrastive learning},
  author={Khosla, Prannay and Teterwak, Piotr and Wang, Chen and Sarna, Aaron and Tian, Yonglong and Isola, Phillip and Maschinot, Aaron and Liu, Ce and Krishnan, Dilip},
  journal={Advances in neural information processing systems},
  volume={33},
  pages={18661--18673},
  year={2020}
}

@article{bepler2021learning,
  title={Learning the protein language: Evolution, structure, and function},
  author={Bepler, Tristan and Berger, Bonnie},
  journal={Cell systems},
  volume={12},
  number={6},
  pages={654--669},
  year={2021},
  publisher={Elsevier}
}

@article{rives2021biological,
  title={Biological structure and function emerge from scaling unsupervised learning to 250 million protein sequences},
  author={Rives, Alexander and Meier, Joshua and Sercu, Tom and Goyal, Siddharth and Lin, Zeming and Liu, Jason and Guo, Demi and Ott, Myle and Zitnick, C Lawrence and Ma, Jerry and others},
  journal={Proceedings of the National Academy of Sciences},
  volume={118},
  number={15},
  pages={e2016239118},
  year={2021},
  publisher={National Academy of Sciences}
}

@article{kim2023functional,
  title={Functional annotation of enzyme-encoding genes using deep learning with transformer layers},
  author={Kim, Gi Bae and Kim, Ji Yeon and Lee, Jong An and Norsigian, Charles J and Palsson, Bernhard O and Lee, Sang Yup},
  journal={Nature Communications},
  volume={14},
  number={1},
  pages={7370},
  year={2023},
  publisher={Nature Publishing Group UK London}
}

@article{wang2025dpfunc,
  title={DPFunc: accurately predicting protein function via deep learning with domain-guided structure information},
  author={Wang, Wenkang and Shuai, Yunyan and Zeng, Min and Fan, Wei and Li, Min},
  journal={Nature communications},
  volume={16},
  number={1},
  pages={70},
  year={2025},
  publisher={Nature Publishing Group UK London}
}

@article{zhu2022integrating,
  title={Integrating unsupervised language model with triplet neural networks for protein gene ontology prediction},
  author={Zhu, Yi-Heng and Zhang, Chengxin and Yu, Dong-Jun and Zhang, Yang},
  journal={PLOS Computational Biology},
  volume={18},
  number={12},
  pages={e1010793},
  year={2022},
  publisher={Public Library of Science San Francisco, CA USA}
}

@article{yang2025dpgok,
  title={DPGOK: A Deep Learning-Based Method for Protein Function Prediction by Fusing GO Knowledge With Protein Features},
  author={Yang, Qiurong and Wang, Wenkang and Fan, Wei and Zheng, Ruiqing and Li, Min},
  journal={IEEE Journal of Biomedical and Health Informatics},
  year={2025},
  publisher={IEEE}
}

@article{kulmanov2023deepgo,
  title={Deepgo-se: Protein function prediction as approximate semantic entailment},
  author={Kulmanov, Maxat and Guzm{\'a}n-Vega, Francisco J and Roggli, Paula Duek and Lane, Lydie and Arold, Stefan T and Hoehndorf, Robert},
  journal={bioRxiv},
  pages={2023--09},
  year={2023},
  publisher={Cold Spring Harbor Laboratory}
}

@article{cao2021tale,
  title={TALE: Transformer-based protein function Annotation with joint sequence--Label Embedding},
  author={Cao, Yue and Shen, Yang},
  journal={Bioinformatics},
  volume={37},
  number={18},
  pages={2825--2833},
  year={2021},
  publisher={Oxford University Press}
}

@article{madden2013blast,
  title={The BLAST sequence analysis tool},
  author={Madden, Thomas},
  journal={The NCBI handbook},
  volume={2},
  number={5},
  pages={425--436},
  year={2013},
  publisher={National Center for Biotechnology Information (US) Bethesda, MD}
}

@article{kulmanov2022deepgozero,
  title={DeepGOZero: improving protein function prediction from sequence and zero-shot learning based on ontology axioms},
  author={Kulmanov, Maxat and Hoehndorf, Robert},
  journal={Bioinformatics},
  volume={38},
  number={Supplement\_1},
  pages={i238--i245},
  year={2022},
  publisher={Oxford University Press}
}

@article{dalkiran2018ecpred,
  title={ECPred: a tool for the prediction of the enzymatic functions of protein sequences based on the EC nomenclature},
  author={Dalkiran, Alperen and Rifaioglu, Ahmet Sureyya and Martin, Maria Jesus and Cetin-Atalay, Rengul and Atalay, Volkan and Do{\u{g}}an, Tunca},
  journal={BMC bioinformatics},
  volume={19},
  number={1},
  pages={334},
  year={2018},
  publisher={Springer}
}

@article{ryu2019deep,
  title={Deep learning enables high-quality and high-throughput prediction of enzyme commission numbers},
  author={Ryu, Jae Yong and Kim, Hyun Uk and Lee, Sang Yup},
  journal={Proceedings of the National Academy of Sciences},
  volume={116},
  number={28},
  pages={13996--14001},
  year={2019},
  publisher={National Academy of Sciences}
}

@article{sanderson2023proteinfer,
  title={ProteInfer, deep neural networks for protein functional inference},
  author={Sanderson, Theo and Bileschi, Maxwell L and Belanger, David and Colwell, Lucy J},
  journal={Elife},
  volume={12},
  pages={e80942},
  year={2023},
  publisher={eLife Sciences Publications Limited}
}

@inproceedings{tan2025venusfactory,
  title={VenusFactory: An Integrated System for Protein Engineering with Data Retrieval and Language Model Fine-Tuning},
  author={Tan, Yang and Liu, Chen and Gao, Jingyuan and Banghao, Wu and Li, Mingchen and Wang, Ruilin and Zhang, Lingrong and Yu, Huiqun and Fan, Guisheng and Hong, Liang and others},
  booktitle={Proceedings of the 63rd Annual Meeting of the Association for Computational Linguistics (Volume 3: System Demonstrations)},
  pages={230--241},
  year={2025}
}

@article{tan2024simple,
  title={Simple, efficient, and scalable structure-aware adapter boosts protein language models},
  author={Tan, Yang and Li, Mingchen and Zhou, Bingxin and Zhong, Bozitao and Zheng, Lirong and Tan, Pan and Zhou, Ziyi and Yu, Huiqun and Fan, Guisheng and Hong, Liang},
  journal={Journal of Chemical Information and Modeling},
  volume={64},
  number={16},
  pages={6338--6349},
  year={2024},
  publisher={ACS Publications}
}

@article{brandes2022proteinbert,
  title={ProteinBERT: a universal deep-learning model of protein sequence and function},
  author={Brandes, Nadav and Ofer, Dan and Peleg, Yam and Rappoport, Nadav and Linial, Michal},
  journal={Bioinformatics},
  volume={38},
  number={8},
  pages={2102--2110},
  year={2022},
  publisher={Oxford University Press}
}

@article{marquet2022embeddings,
  title={Embeddings from protein language models predict conservation and variant effects},
  author={Marquet, C{\'e}line and Heinzinger, Michael and Olenyi, Tobias and Dallago, Christian and Erckert, Kyra and Bernhofer, Michael and Nechaev, Dmitrii and Rost, Burkhard},
  journal={Human genetics},
  volume={141},
  number={10},
  pages={1629--1647},
  year={2022},
  publisher={Springer}
}

@article{bateman2000pfam,
  title={The Pfam protein families database},
  author={Bateman, Alex and Birney, Ewan and Durbin, Richard and Eddy, Sean R and Howe, Kevin L and Sonnhammer, Erik LL},
  journal={Nucleic acids research},
  volume={28},
  number={1},
  pages={263--266},
  year={2000},
  publisher={Oxford University Press}
}

@article{uniprot2021uniprot,
  title={UniProt: the universal protein knowledgebase in 2021},
  author={UniProt Consortium},
  journal={Nucleic acids research},
  volume={49},
  number={D1},
  pages={D480--D489},
  year={2021},
  publisher={Oxford University Press}
}

@article{li2018deepre,
  title={DEEPre: sequence-based enzyme EC number prediction by deep learning},
  author={Li, Yu and Wang, Sheng and Umarov, Ramzan and Xie, Bingqing and Fan, Ming and Li, Lihua and Gao, Xin},
  journal={Bioinformatics},
  volume={34},
  number={5},
  pages={760--769},
  year={2018},
  publisher={Oxford University Press}
}

@article{altschul1990basic,
  title={Basic local alignment search tool},
  author={Altschul, Stephen F and Gish, Warren and Miller, Webb and Myers, Eugene W and Lipman, David J},
  journal={Journal of molecular biology},
  volume={215},
  number={3},
  pages={403--410},
  year={1990},
  publisher={Elsevier}
}

\newpage
\appendix
\section{Related Work}

\subsection{Dynamics of Representation Learning and Geometric Diagnostics}
Recent work suggests that knowledge acquisition in large language models is not a smooth, uniform process, but often happens in stages: models can show long performance plateaus while internal mechanisms reorganize, followed by later gains that reflect a different internal organization ~\cite{power2022grokking,zucchet2025language}.
Related findings indicate that networks tend to learn simpler statistical regularities before higher-order structure~\cite{refinetti2023neural}, which is consistent with observations that representation geometry and spectral properties can evolve non-monotonically over training~\cite{li2025tracing}.
These observations motivate diagnostics that track representation structure directly, rather than relying only on task performance. Recent works have introduced covariance-spectrum diagnostics that quantify directional dominance and effective dimensionality, such as RankMe~\cite{garrido2023rankme} and spectral decay measures~\cite{agrawal2022alreq}. Such tools are particularly useful for post-training settings, where objectives can substantially reshape representation space.

\subsection{Post-training and Representation Shaping}
Post-training is widely used to adapt pretrained models through supervised fine-tuning, preference optimization (e.g., RLHF or direct preference objectives such as DPO), and related reward-based variants~\cite{ouyang2022training, rafailov2023direct, shao2024deepseekmath}, often combined with parameter-efficient updates such as adapters or LoRA~\cite{houlsby2019parameter, hu2022lora}.
Beyond behavioral alignment, many post-training objectives can also be viewed as reshaping representation space by enforcing invariances and improving separability, as studied in contrastive learning and supervised contrastive formulations~\cite{chen2020simple, khosla2020supervised}.
We adopt this perspective to characterize post-training dynamics directly through spectral diagnostics, linking geometric shifts to downstream function prediction performance.

\subsection{Protein Language Models and Function Prediction}
Protein language models pretrained on large-scale protein sequence data have become a foundation for protein modeling, enabling transfer to structure and function prediction via task-specific fine-tuning ~\cite{rives2021biological,elnaggar2021prottrans,lin2023evolutionary,bepler2021learning}.
In contrast to LLMs, PLMs are often deployed directly after pretraining, and post-training is less standardized.
For function prediction, a common paradigm is to treat PLM embeddings as general-purpose features and train a supervised classifier on curated labels such as enzyme commission or gene ontology~\cite{zhu2022integrating,yu2023enzyme,kulmanov2023deepgo,tan2024simple,yang2025interpretable,wang2025dpfunc,yang2025dpgok}.
Complementary to this paradigm, non-PLM approaches targets function prediction with sequence-similarity or task-specific architectures without relying on pretrained PLMs~\cite{madden2013blast,cao2021tale,kulmanov2022deepgozero,cao2021tale}.
Existing adaptation practice is dominated by supervised fine-tuning and continued pretraining, while comparatively fewer works explicitly target representation shaping using sequence-derived relational supervision at scale.
This gap motivates our approach: we post-train PLM representations with a contrastive objective grounded in Pfam family co-membership, which provides a invariance signal that is correlated with function yet distinct from task labels.

\section{Impact Statement}

This work aims to improve sequence-based protein function prediction by reshaping pretrained protein language model representations with homology-guided post-training. Potential positive impacts include more accurate computational annotation of protein functions, improved prioritization of candidate enzymes or proteins for downstream experimental validation, and more efficient reuse of pretrained protein encoders in biological discovery pipelines. Because SAVR uses only sequence-derived Pfam co-membership during post-training and does not require task-specific EC/GO labels, it may also reduce the amount of labeled data needed for adapting PLMs to protein-level function prediction tasks.

The main risks are those shared by computational protein function prediction methods. Incorrect predictions may mislead biological interpretation if used without experimental validation, and improved function prediction tools could potentially be incorporated into workflows for engineering proteins with harmful properties. SAVR is not a generative protein design method and does not directly propose new sequences, structures, or experimental protocols. Nevertheless, we recommend that predictions from SAVR-based systems be treated as computational hypotheses and validated by appropriate biological experiments, especially in safety-sensitive applications. The method is intended to support responsible protein annotation and representation learning research rather than autonomous biological design.

\section{Raw Geometry Metrics}
\label{appendix:raw}
Table~\ref{tab:geometry-raw} reports representation-geometry statistics computed on \emph{raw} pooled embeddings \emph{without} row-wise $\ell_2$ normalization, included in the appendix as an unnormalized reference. Across both backbones, ECT induces a consistent spectral reshaping pattern: compared to the vanilla representations, variance becomes less dominated by a few principal directions and more distributed across dimensions, with the piecewise spectral decay exhibiting reduced head dominance and a stronger long-tail contribution. The power-law exponents are fitted over three index ranges, with the number of eigenvalues used in each fit given by: for \textbf{ESM-2 / ESM-ECT} ($d{=}1280$), head $[1,3]$ ($n{=}3$), mid $[4,54]$ ($n{=}51$), and tail $[55,1280]$ ($n{=}1226$); for \textbf{ProtBert} ($d{=}1024$), head $[1,1]$ ($n{=}1$), mid $[2,7]$ ($n{=}6$), and tail $[8,1024]$ ($n{=}1017$); and for \textbf{PB-ECT} ($d{=}1024$), head $[1,8]$ ($n{=}8$), mid $[9,83]$ ($n{=}75$), and tail $[84,1024]$ ($n{=}941$). Note that the vanilla ProtBert head segment contains only a single eigenvalue ($n{=}1$), making $\alpha_h$ ill-defined (NaN in the log); we therefore omit it in the table.

\begin{table}[t]
  \caption{Representation geometry analysis. Geometry metrics on pooled embeddings (RankMe, Top-10 variance ratio, and spectral decay exponents; without row-wise $\ell_2$ normalization)}
  \label{tab:geometry-raw}
  \centering
  \setlength{\tabcolsep}{5pt}
  \renewcommand{\arraystretch}{1.1}
  \begin{small}
    \begin{sc}
      \begin{tabular}{l@{\hspace{3pt}}c@{\hspace{4pt}}cccc}
        \toprule
        Method & RankMe$\uparrow$ & Top-10$\downarrow$ & $\alpha_{h}$ & $\alpha_{m}$ & $\alpha_{t}$\\
        \midrule
        ESM-2 & 53.1 & 0.574 & 1.617 & 1.067 & 2.363 \\
        ESM-ECT & \textbf{93.4} & \textbf{0.441} & 1.107 & 1.036 & 2.727 \\
        \midrule
        ProtBert & 9.93 & 0.835 & - & 1.799 & 1.528 \\
        PB-ECT & 91.0 & 0.456 & 0.991 & 1.091 & 2.459 \\
        \bottomrule
      \end{tabular}
    \end{sc}
  \end{small}
  \vskip -0.08in
\end{table}

\section{Row-wise \texorpdfstring{$\ell_2$}{L2} normalization}
\label{appendix:normalization}
Given pooled sequence representations $\{\mathbf{z}_i\}_{i=1}^N$ with $\mathbf{z}_i\in\mathbb{R}^d$, we apply row-wise $\ell_2$ normalization to remove per-sequence scale variation:
\begin{equation}
\hat{\mathbf{z}}_i \;=\; \frac{\mathbf{z}_i}{\|\mathbf{z}_i\|_2 + \epsilon}, \qquad i=1,\ldots,N,
\label{eq:rowl2}
\end{equation}
where $\epsilon>0$ is a small constant for numerical stability.
All spectrum-based diagnostics reported under the ``row-wise $\ell_2$ normalized'' setting are computed from the empirical covariance of the centered, normalized representations.
Specifically, letting $\hat{\mathbf{Z}}\in\mathbb{R}^{N\times d}$ denote the matrix whose $i$-th row is $\hat{\mathbf{z}}_i$, we center
\begin{equation}
\tilde{\mathbf{Z}} \;=\; \hat{\mathbf{Z}} - \mathbf{1}\,\hat{\boldsymbol{\mu}}^\top,
\qquad
\hat{\boldsymbol{\mu}} \;=\; \frac{1}{N}\sum_{i=1}^N \hat{\mathbf{z}}_i,
\label{eq:center}
\end{equation}
and compute the covariance
\begin{equation}
\boldsymbol{\Sigma} \;=\; \frac{1}{N-1}\tilde{\mathbf{Z}}^\top \tilde{\mathbf{Z}} \in \mathbb{R}^{d\times d}.
\label{eq:cov}
\end{equation}
Row-wise $\ell_2$ normalization ensures that each sequence contributes with comparable magnitude (i.e., $\|\hat{\mathbf{z}}_i\|_2 \approx 1$), so the resulting spectrum primarily reflects directional geometry on the unit hypersphere rather than variation in embedding norms.

\section{Full Positive-Pair Construction Control Results}
\label{appendix:contrastive_ablation}

\begin{table}[t]
\centering
\small
\setlength{\tabcolsep}{6pt}
\renewcommand{\arraystretch}{1.12}
\caption{Comparison of different contrastive post-training strategies.}
\label{tab:contrastive_ablation_full}
\begin{tabular}{lccccc}
\toprule
Method & RankMe$\uparrow$ & Top-10$\downarrow$ & $\alpha_h$ & EC New F1 & EC Price F1 \\
\midrule
Random CL        & 10.9 (collapse) & 0.865 & --    & 0.317 & 0.217 \\
SI $\geq$30\% CL & 94.0            & 0.446 & 1.080 & 0.582 & 0.562 \\
SAVR             & \textbf{122.5}  & \textbf{0.372} & \textbf{0.749} & \textbf{0.616} & \textbf{0.607} \\
\bottomrule
\end{tabular}
\end{table}

Table~\ref{tab:contrastive_ablation_full} reports the full control results for positive-pair construction. Random positives lead to representation collapse and poor EC performance, while sequence-identity positives provide a stronger baseline but remain below SAVR. SAVR achieves the best geometry and downstream EC results, supporting that Pfam co-membership provides a more effective positive-pairing signal than random or fixed sequence-identity-based grouping.

\section{Additional Leakage Risk Analysis}
\label{app:sec:leakage}

We evaluate whether SAVR introduces an additional leakage channel beyond the overlap structure already present in the downstream EC/GO benchmarks. The answer is no under both sequence-identity and Pfam-family overlap analyses. SAVR never uses EC or GO labels during post-training; Pfam annotations are used only to construct positive pairs, and downstream labels are observed only when training the task-specific prediction head.

First, we compare each test set against two reference sets: the SAVR post-training corpus and the benchmark's own downstream training split. Using MMseqs2 best-hit sequence identity, we find that test proteins are generally more similar to the benchmark training split than to the SAVR post-training corpus (Table~\ref{tab:leakage_mmseqs}). Thus, the post-training corpus is not a stronger sequence-overlap source than the training data already used by the accepted benchmark protocols.

\begin{table}[t]
\centering
\small
\setlength{\tabcolsep}{6pt}
\renewcommand{\arraystretch}{1.12}
\caption{Best-hit sequence-identity overlap between downstream test sets and two reference sets: the SAVR post-training corpus and the corresponding benchmark training split. Values report the percentage of test proteins with best-hit identity above each threshold.}
\label{tab:leakage_mmseqs}
\begin{tabular}{llccc}
\toprule
Test set & Reference & $\geq$50\% & $\geq$70\% & $\geq$90\% \\
\midrule
EC New-392   & SAVR  & 38.0 & 12.0 & 4.1 \\
             & EC train & 53.8 & 25.0 & 9.9 \\
EC Price-149 & SAVR  & 36.9 & 8.7  & 1.3 \\
             & EC train & 30.9 & 18.1 & 1.3 \\
GO-BP        & SAVR  & 41.8 & 18.7 & 7.0 \\
             & GO-BP train & 50.7 & 34.2 & 21.0 \\
GO-CC        & SAVR  & 44.9 & 22.4 & 9.3 \\
             & GO-CC train & 53.4 & 37.7 & 25.9 \\
GO-MF        & SAVR  & 35.6 & 16.5 & 7.0 \\
             & GO-MF train & 50.7 & 36.2 & 24.5 \\
\bottomrule
\end{tabular}
\end{table}

Second, we analyze Pfam-family overlap. A test protein is counted as overlapping with a reference set if at least one of its Pfam families appears in that reference set. Pfam overlap is already nearly saturated in the benchmark train/test splits themselves (Table~\ref{tab:leakage_pfam_overlap}). The SAVR post-training corpus shows comparable overlap, but this is not a special condition introduced by SAVR; it reflects the family-level redundancy inherent to standard EC/GO function-prediction benchmarks.

\begin{table}[t]
\centering
\small
\setlength{\tabcolsep}{7pt}
\renewcommand{\arraystretch}{1.12}
\caption{Pfam-family overlap between downstream test proteins and two reference sets. A test protein is counted as overlapping if at least one of its Pfam families appears in the reference set.}
\label{tab:leakage_pfam_overlap}
\begin{tabular}{lcc}
\toprule
Test set & SAVR post-training overlap & Benchmark train overlap \\
\midrule
EC New   & 99.5\% & 100.0\% \\
EC Price & 100.0\% & 100.0\% \\
GO-BP    & 94.8\% & 90.4\% \\
GO-CC    & 95.0\% & 90.2\% \\
GO-MF    & 95.2\% & 87.9\% \\
\bottomrule
\end{tabular}
\end{table}

These results rule out the interpretation that SAVR gains arise from a uniquely privileged overlap between the post-training corpus and the downstream test sets. At the sequence level, SAVR has no stronger test-set overlap than the benchmark training splits. At the Pfam-family level, overlap is already a pervasive property of the benchmarks themselves. Therefore, the observed improvements are not explained by an additional leakage channel introduced by SAVR, but by task-agnostic post-training with a biologically meaningful grouping signal.

\section{Implementation Details}
\label{appendix:implementation_details}

\subsection{Training data and batch construction}
We train using a Pfam-annotated protein dataset organized by Pfam families. Each protein sequence is paired with a \emph{set} of Pfam domain IDs (allowing multi-domain proteins).
Mini-batches are constructed with two hyperparameters $(P, K)$: for each batch, we sample $P$ Pfam families and then sample $K$ sequences per family, yielding an effective batch size of $B=P\times K$ (in our main setting, $P{=}8$, $K{=}2$, so $B{=}16$). We run data parallel on 4 NVIDIA V100 GPUs, and the mini-batch on each GPUs are gathered to formed the global batch with equivalent batch size of 64.
Sequences are truncated to a maximum length of 1024 amino acids; tokenization uses standard model tokenizers with padding and truncation enabled.

\subsection{Backbone and parameter-efficient tuning}
We use \texttt{facebook/esm2\_t33\_650M\_UR50D} and \texttt{Rostlab/prot\_bert} as the backbone encoder and apply LoRA for parameter-efficient adaptation.
LoRA is injected into the attention projection modules \texttt{query} and \texttt{value}.
We use LoRA rank $r{=}16$, scaling $\alpha{=}16$, and LoRA dropout $0.05$, with no bias parameters adapted.
To reduce activation memory and stabilize training, gradient checkpointing is enabled and the model cache is disabled.

\subsection{Sequence representation and projection head}
Given a tokenized sequence, we extract the final-layer hidden states and compute a mean-pooled sequence embedding by averaging over non-padding tokens while excluding special tokens (e.g., BOS/CLS and EOS when applicable).
A two-layer MLP projection head maps the pooled embedding to a 512-dimensional contrastive space:
\[
\mathrm{MLP}(x)=W_2(\mathrm{Dropout}(\mathrm{LN}(\mathrm{ReLU}(W_1 x)))) \in \mathbb{R}^{512},
\]
with hidden dimension 1024 and dropout 0.1.
The projected feature is $\ell_2$-normalized before similarity computation.

\paragraph{Distributed global negatives.}
In distributed training, we compute the denominator over features gathered across all processes, while keeping gradients only for local anchors. This provides a larger and more diverse negative set without duplicating loss computation across ranks.

\subsection{Optimization and training schedule}
We use AdamW with two parameter groups:
(1) LoRA-adapted backbone parameters, learning rate $2\times 10^{-4}$, weight decay $0.01$;
(2) projection head parameters, learning rate $10^{-3}$, weight decay $0.01$ for weight matrices.
AdamW betas are set to $(0.9, 0.95)$.

We train for 5 epochs with automatic mixed precision.
The learning rate follows a cosine decay schedule with linear warmup over the first 5\% of total optimization steps.

\section{Experiment Details}
\label{appendix:exp_implementation}

\subsection{MLP Classifier}
\label{sec:app:mlp}
\paragraph{Mean-pooling classifier.}
For all EC and GO tasks, we use the same lightweight mean-pooling classifier on top of frozen PLM token embeddings.
Given token-level representations $\mathbf{H}\in\mathbb{R}^{L\times d}$ from the PLM and an attention mask $m\in\{0,1\}^{L}$, we compute a mask-aware mean pooled sequence embedding
\begin{equation}
\mathbf{z}=\frac{\sum_{t=1}^{L} m_t \mathbf{H}_t}{\sum_{t=1}^{L} m_t + \epsilon}\in\mathbb{R}^{d},
\end{equation}
where $\epsilon$ is a small constant for numerical stability.

The pooled embedding is then fed into a two-layer MLP with a residual-width hidden layer:
\begin{equation}
\label{appendix:mlp_formular}
\mathbf{u}=\mathrm{Dropout}(\mathbf{z}),\quad
\mathbf{v}=\mathrm{GELU}(\mathbf{W}_1 \mathbf{u}+\mathbf{b}_1),\quad
\hat{\mathbf{y}}=\mathbf{W}_2\,\mathrm{Dropout}(\mathbf{v})+\mathbf{b}_2,
\end{equation}
where $\mathbf{W}_1\in\mathbb{R}^{d\times d}$ and $\mathbf{W}_2\in\mathbb{R}^{C\times d}$, and $C$ is the number of EC labels (classes). We use the final-layer hidden size of the backbone PLM as the representation dimension d. Specifically, d=\texttt{hidden\_size} of the pretrained encoder (d=1280 for ESM2-650M and d=1024 for ProtBERT). We use dropout rate $p=0.1$ (unless otherwise specified) and GELU as activation function.

\paragraph{Attention-pooling classifier.}
As an alternative classifier, we replace mean pooling with a learned attention pooling module while keeping the same two-layer MLP classifier as in the mean-pooling head.
Given token representations $\mathbf{H}\in\mathbb{R}^{L\times d}$ and a binary attention mask $m\in{0,1}^{L}$, we compute a scalar score for each token using a two-layer MLP:
\begin{equation}
s_t = \mathbf{w}_2^\top \tanh(\mathbf{W}_a \mathbf{H}_t + \mathbf{b}_a) + b_2,
\end{equation}
where $\mathbf{W}a\in\mathbb{R}^{512\times d}$, $\mathbf{w}2\in\mathbb{R}^{512}$, $\mathbf{b}a\in\mathbb{R}^{512}$, and $b_2\in\mathbb{R}$.
We apply mask-aware normalization by setting $s_t=-\infty$ for masked tokens ($m_t=0$) and computing attention weights
\begin{equation}
a_t = \frac{\exp(s_t)}{\sum{t'=1}^{L} \exp(s{t'})}.
\end{equation}
The pooled sequence embedding is then the weighted sum
\begin{equation}
\mathbf{z}=\sum{t=1}^{L} a_t \mathbf{H}_t \in \mathbb{R}^{d}.
\end{equation}
Finally, we feed $\mathbf{z}$ into the same MLP head as Eq.~\ref{appendix:mlp_formular}, with dropout and GELU, producing logits $\hat{\mathbf{y}}\in\mathbb{R}^{C}$.
We use the same representation dimension $d$ (the backbone hidden size), the same dropout rate $p$, and the same optimization setup as the mean-pooling classifier.

\subsection{Optimization and training schedule.}
We follow prior work for the optimizer and training schedules on each benchmark. On VenusFactory, we use AdamW with a learning rate of $5\times10^{-4}$ and train for up to 100 epochs, with early stopping (patience = 10). We use a maximum batch length of 12{,}000 tokens and set gradient accumulation steps to 8. For GO prediction, we follow DPGOK: we use AdamW with a learning rate of $5\times10^{-4}$ and train for 25/20/30 epochs for MFO/BPO/CCO, respectively. For EC prediction, we follow the ProtDETR setting and use AdamW with a learning rate of $1\times10^{-4}$ for 50 epochs.

\section{Identity-binned (Homology) Evaluation}

To diagnose whether the gains come from improved remote-homology generalization rather than near-duplicate interpolation, we stratify the test set by the maximum sequence identity between each test protein and the downstream training set. Concretely, we compute $\max_{x \in \mathcal{D}_{\mathrm{train}}}\mathrm{Id}(x_{\mathrm{test}}, x)$ and report Macro-F1 within each identity bin. To reduce the influence of classifier capacity and focus on representation quality, we use a single linear probe for this analysis.

\begin{wrapfigure}{r}{0.5\textwidth}
  \centering
  \vspace{-1.0em}
  \includegraphics[width=\linewidth]{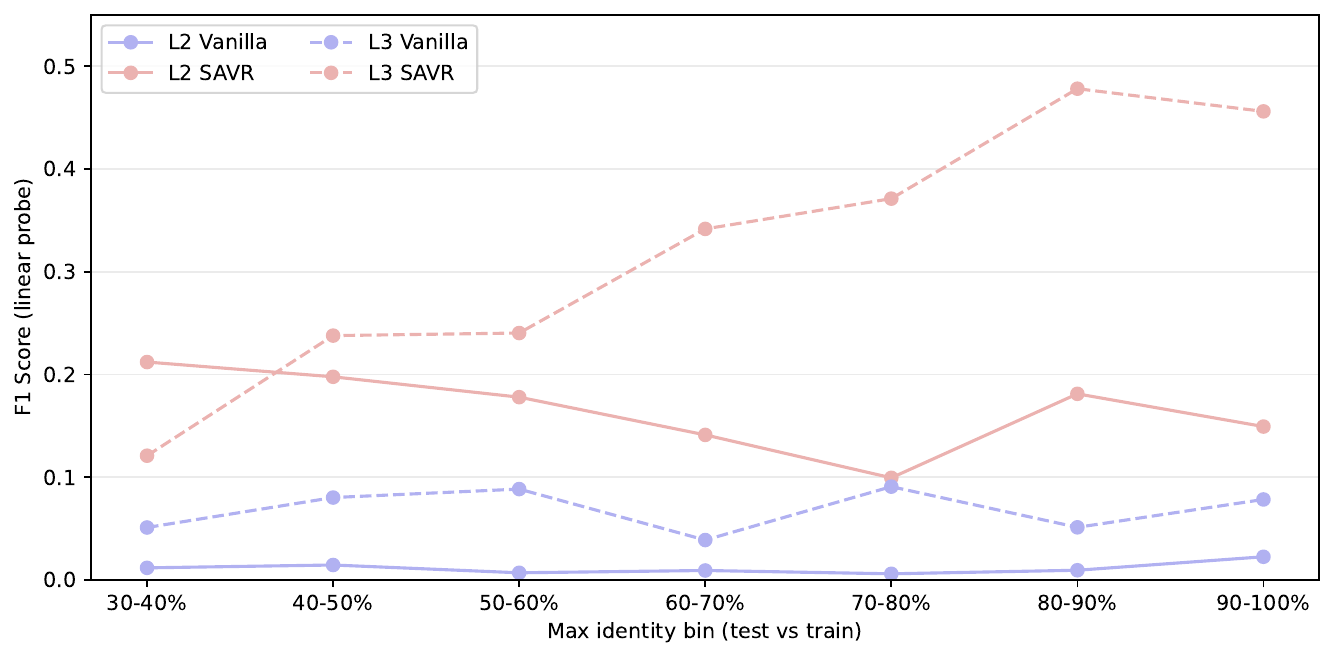}
  \vspace{-0.8em}
  \caption{EC Level-2 and Level-3 macro-F1 (linear probe) across max train--test sequence-identity bins, comparing Vanilla vs SAVR.}
  \label{fig:ec23_id_bin}
\end{wrapfigure}

Fig.~\ref{fig:ec23_id_bin} compares the vanilla backbone and SAVR for EC prediction at Level-2 and Level-3. Across bins, SAVR consistently outperforms the vanilla model, including in low-identity regimes (30--60\%), where performance reflects remote-homology generalization rather than similarity-based matching. Moreover, the advantage persists across EC granularities: while Level-2 is substantially more challenging and the vanilla representations yield near-zero Macro-F1 in several bins, SAVR maintains a markedly stronger linear readout signal. These results suggest that SAVR improves function-relevant structure in the embedding space in a way that generalizes beyond close homologs.



\end{document}